\pdfoutput=1

\documentclass[11pt]{article}

\usepackage[final]{acl}

\usepackage{times}
\usepackage{latexsym}

\usepackage[T1]{fontenc}

\usepackage[utf8]{inputenc}

\usepackage{microtype}

\usepackage{inconsolata}

\usepackage{graphicx}

\usepackage{amsmath} 
\usepackage{booktabs}
\usepackage{pifont}
\usepackage{makecell}

\usepackage[switch]{lineno}

\usepackage{subcaption}

\title{ResearchBench: Benchmarking LLMs in Scientific Discovery \\via Inspiration-Based Task Decomposition}

\author{
    Yujie Liu\textsuperscript{\rm 1,2}$^*$,
    Zonglin Yang\textsuperscript{\rm 3,2}$^*$,
    Tong Xie\textsuperscript{\rm 4},
    Jinjie Ni\textsuperscript{\rm 5},
    Ben Gao\textsuperscript{\rm 6,2}, \\
    \textbf{Yuqiang Li}\textsuperscript{\rm 2},
    \textbf{Shixiang Tang}\textsuperscript{\rm 2},
    \textbf{Wanli Ouyang}\textsuperscript{\rm 2}$^\dagger$,
    \textbf{Erik Cambria}\textsuperscript{\rm 3}$^\dagger$,
    \textbf{Dongzhan Zhou}\textsuperscript{\rm 2}$^\dagger$\\
    \textsuperscript{\rm 1} {\small Fudan University}
    \textsuperscript{\rm 2} {\small Shanghai Artificial Intelligence Laboratory}
    \textsuperscript{\rm 3} {\small Nanyang Technological University}\\
    \textsuperscript{\rm 4} {\small University of New South Wales}
    \textsuperscript{\rm 5} {\small National University of Singapore}
    \textsuperscript{\rm 6} {\small Wuhan University}\\
    {\tt \small liuyujie.cs@gmail.com},
    {\tt \small \{zonglin001,cambria\}@ntu.edu.sg},
    {\tt \small zhoudongzhan@pjlab.org.cn} \\
}

\begin{document}
\maketitle
\begin{abstract}

Large language models (LLMs) have shown potential in assisting scientific research, yet their ability to discover high-quality research hypotheses remains unexamined due to the lack of a dedicated benchmark. To address this gap, we introduce the first large-scale benchmark for evaluating LLMs on a sufficient set of scientific discovery sub-tasks—inspiration retrieval, hypothesis composition, and hypothesis ranking—where sufficient means that perfectly solving these sub-tasks perfectly solves the overall discovery task. We develop an automated LLM-based framework that extracts critical components—research questions, background surveys, inspirations, and hypotheses—from papers across 12 disciplines, with expert validation confirming its accuracy. To prevent data contamination, we focus exclusively on publications from 2024 onward, ensuring minimal overlap with LLM pretraining data; our automated framework further enables automatic extraction of even more recent papers as LLM pretraining cutoffs advance, supporting scalable and contamination-free automatic renewal of this discovery benchmark. Our evaluation shows that, across disciplines, LLMs excel at inspiration retrieval—an out-of-distribution task—suggesting their ability to surface novel knowledge associations.
\let\thefootnote\relax\footnote{$^*$Both authors contributed equally to this work.}\let\thefootnote\relax\footnote{$^\dagger$Corresponding author.}

\end{abstract}

\section{Introduction}

Large language models~(LLMs) have shown potential to assist scientists' research as copilots~\citep{survey}.
One of the most challenging copilot tasks is helping scientists discover valid new research hypotheses, where a typical setting provides only a research question and a background survey as input.
Understanding how LLMs perform on this task is crucial for selecting appropriate models and evaluating how different training strategies influence their effectiveness in scientific discovery.
However, although some efforts benchmark LLMs' performance on general tasks, such as Chatbot Arena~\citep{chatbotarena} and MixEval~\citep{ni2024mixeval}, it still lacks understanding of each LLM's scientific discovery ability.

A primary reason for this gap is the limited understanding of the scientific discovery process—specifically, how research hypotheses are formulated. Recently, \citet{msc} decomposed this process into a sufficient set of sub-tasks--meaning that perfectly solving these sub-tasks perfectly solves the overall discovery task: (1) retrieving inspirations based on the research question, (2) mixing the research background with retrieved inspirations to compose hypotheses, and (3) ranking the composed hypotheses. This decomposition is viable under the fundamental assumption that most hypotheses originate from a research background combined with inspirations.

This fundamental assumption is supported by cognitive science findings that \textit{creative ideas often result from the cohesive association of two~(or more) seemingly unrelated pieces of knowledge}~\citep{koestler1964act,benedek2012associative,lee2024empirical}.
The cognitive science findings are discipline-independent and widely applicable.
For example, the proposal of backpropagation is a research hypothesis. In this case, the research background is about multi-layer logistic regression, and the inspiration is the chain rule in calculus.


\begin{table*}
  \centering
  \resizebox{1.8\columnwidth}{!}{
  \begin{tabular}{c|cccccccccccc|c}
    \toprule
    \textbf{Discipline} & Cell & Chem & ETS & MS & Phys & EGS & EVS & BL & BS & Law & Math & AT & Overall \\
    \midrule
    \textbf{Paper Number} & 152 & 113 & 114 & 116 & 132 & 117 & 116 & 115 & 115 & 97 & 113 & 86 & 1386 \\
    \bottomrule
  \end{tabular}}
  \caption{Disciplines and paper number distribution. Chem=Chemistry, ETS=Earth Science, MS=Material Science, Phys=Physics, EGS=Energy Science, EVS=Environmental Science, BL=Biology, BS=Business, AT=Astronomy.}
  \label{tab:disciplines_paper_number}
\end{table*}

This research aims at filling the research gap, by providing a benchmark specifically designed to evaluate LLM's performance in terms of the three decomposed tasks of scientific discovery. 
This benchmark covers 12 disciplines, selecting papers published on top venues such as Nature, Science, or journals of a similar level; the discipline distribution and paper counts are summarized in Table~\ref{tab:disciplines_paper_number}.

To construct the benchmark, we download 1386 papers and develop an automated LLM-based agentic framework to analyze each paper into research question, background survey, inspirations, and main hypothesis. 

We invited five experts in Physics, Chemistry, Astronomy, and Materials Science disciplines to check whether the decomposition is accurate. 
Among the randomly sampled 62 papers checked by the experts, the decomposition accuracy was 91.9\% considering only major issues and 82.3\% when including both major and minor issues.
It shows that the automated framework can accurately extract these components from a paper.

To prevent data contamination, we select only papers published from 2024 onward.
The advantage of our LLM agentic framework for the extraction is that as the LLM's pretraining data cutoff date moves forward, the framework can automatically extract more recent papers to avoid overlapping. 

Based on the benchmark, we systematically compare popular LLMs across the three decomposed tasks. We find that current LLMs perform well in retrieving inspirations across disciplines, despite the inclusion of carefully crafted challenging negative inspiration examples.
For example, when we ask GPT-4o to select the top 4\% of inspiration candidates, the probability that a ground truth inspiration will be included in the top 4\% is 45.7\%.
It is surprising because the inspiration retrieval task is actually an out-of-distribution~(OOD) task since inspiration is supposed to be \emph{not known} to be related to the research question but in fact \emph{can assist} it. 
Otherwise, the resulting hypothesis won't be novel. 
In addition, we find that LLMs also have a good performance on the hypothesis composition and hypothesis ranking task. 
This suggests that LLMs could be leveraged as research hypothesis mines, where higher-performing LLMs on the three fundamental tasks of scientific discovery act as richer mines, and more inference compute corresponds to more miners.

Overall, the contributions of this paper are:
\begin{itemize}
    \item We introduce the first large-scale benchmark for evaluating LLMs’ capabilities in scientific discovery, focusing on a theoretically grounded, sufficient set of sub-tasks: inspiration retrieval, hypothesis composition, and hypothesis ranking.
    \item We develop an automated agentic framework to extract essential components—research questions, surveys, inspirations, and hypotheses—from scientific papers with high accuracy, enabling the scalable and contamination-resistant construction of benchmarks.
    \item We conduct a comprehensive analysis of LLM performance using our benchmark, presenting the first large-scale study on out-of-distribution (OOD) inspiration retrieval. Our findings demonstrate that LLMs can effectively retrieve inspirations beyond established associations, positioning LLMs as “research hypothesis mines” capable of generating novel scientific insights at scale with minimal human involvement.
\end{itemize}

\section{Related Work}

\subsection{Inspiration-Based Scientific Discovery}
MOOSE~\citep{moose} and MOOSE-Chem~\citep{msc} find that most hypotheses in social science, chemistry, and materials science can be seen as emerging from a composition of the research background and several inspirations. 
\citet{scimon} also study how retrieved concepts aid hypothesis generation.
In addition, many researchers have the belief that ``An idea is nothing more nor less than a new combination of old elements''~\citep{young1975technique,kumar,garikaparthi2025mir}.
\citet{msc} further decompose the seemingly intractable scientific discovery task into a sufficient set of sub-tasks: inspiration retrieval, hypothesis composition, and hypothesis ranking: in a sense that perfectly solving the three sub-tasks should lead to perfectly solving the overall scientific discovery task. 
This sufficiency property is discussed in \S~\ref{sec:preliminary}.

\subsection{Benchmarking LLMs}
Most existing benchmarks assess LLMs' general intelligence~\citep{chatbotarena,ni2024mixeval}. IdeaBench~\citep{ideabench} targets biomedical idea generation without evaluating the full set of scientific discovery sub-tasks; it focuses solely on generating hypotheses from background knowledge, omitting inspiration retrieval and integration, uses rule-based extraction (less accurate than our LLM-based agentic framework), and is limited to the biomedical domain (unlike our 12 disciplines). DiscoveryBench~\citep{DiscoveryBench} and ScienceAgentBench~\citep{ScienceAgentBench} extract specific discovery-relevant tasks~(e.g., writing code) from 20 and 44 papers, respectively, but do not analyze the fundamental decomposition of scientific discovery. PaperBench~\citep{starace2025paperbench} emphasizes experimental implementation over hypothesis discovery, while LLM-SRBench~\citep{shojaee2025llm} focuses on symbolic regression without generalizing across disciplines.


\section{Benchmark Construction}

\subsection{Theoretical Foundation}
\label{sec:preliminary}

\citet{msc} propose a fundamental assumption that a majority of chemistry and materials science hypotheses can originate from a research background and several inspirations.
Here research background refers to a research question and/or a background survey; inspiration is a piece of knowledge, and it can be also represented by a research paper that discusses this piece of knowledge.
They propose this assumption based on extensive discussions with domain experts, and the cognitive science findings that \textit{creative ideas often result from the cohesive association of two (or more) seemingly unrelated pieces of knowledge}~\citep{koestler1964act,benedek2012associative,lee2024empirical}.
Denoting background knowledge as $b$, inspiration knowledge as $i$, and hypothesis as $h$, this assumption can be represented in Equation~\ref{eq:assumption}:

\begin{equation}
    h = f(b, i_1, \ldots, i_k)
    \label{eq:assumption}
\end{equation}



Based on this assumption, \citet{msc} decompose hypothesis formulation into a sufficient set of sub-tasks: (1) retrieving inspirations based on the research background, (2) mixing the background information with retrieved inspirations to compose hypotheses, and (3) ranking the composed hypotheses. This process is represented in Equations~\ref{eq:decomposition} and~\ref{eq:ranking}, where $I$ denotes the full literature corpus for inspiration retrieval; specifically, Equation~\ref{eq:decomposition} approximates \(P(h|b)\) as an iterative process of inspiration retrieval followed by hypothesis composition, equating ``hypothesis discovery'' to ``inspiration retrieval \(\to\) hypothesis composition''~(iteratively), with the key distinction that this generates multiple candidate hypotheses, whereas true discovery yields only one or a few. Ranking thus completes the decomposition (via Equation~\ref{eq:ranking}), rendering ``hypothesis discovery'' equivalent to ``inspiration retrieval \(\to\) hypothesis composition \(\to\) ranking,'' such that these three sub-tasks form a sufficient set for scientific discovery—in the sense that perfectly solving them perfectly solves the overall discovery task.

\begin{equation}
    \fontsize{10pt}{11pt}\selectfont
     P(h|b) \approx \prod_{j=1}^{k} P(i_j|b, h_{j-1}, I) \cdot P(h_j|b, h_{j-1}, i_j)
    \label{eq:decomposition}
\end{equation}

\begin{equation}
    \fontsize{9.5pt}{11pt}\selectfont
    H = \{ h_1, h_2, \ldots, h_n \mid \hat{R}(h_i)>\hat{R}(h_{i+1}) \text{ for all } i \}
\label{eq:ranking}
\end{equation}

The cognitive science findings are not limited to any single discipline.
For example, \citet{moose} shows that this assumption can also be leveraged in social science disciplines to generate high-quality research hypothesis.
After extensive discussions with researchers in other disciplines such as Physics, Biology, Earth Science, Astronomy, and Math, we find that this assumption is largely and widely true across disciplines.

Based on this observation, we construct this benchmark collecting papers across 12 disciplines in top-ranked venues, develop an agentic framework to automatically extract each paper's research background, inspirations, and hypothesis~(\S~\ref{subsec:agentic_framework}), discuss how negative inspiration papers are selected to evaluate LLMs' performance on inspiration retrieval~(\S~\ref{subsec:negative_insp_select}), and present expert evaluation on the extracted information to illustrate the quality of the benchmark~(\S~\ref{subsec:expert_evaluation}).

Unlike directly assigning a score to each hypothesis and ranked the hypotheses based on their scores~(Equation~\ref{eq:ranking}), this benchmark adopts a pairwise evaluation to rank~(Equation~\ref{eq:ranking_pairwise}), since pairwise evaluation is widely reported as more robust and reliable~\citep{si2024can}.
$R(h_i, h_{i+1})=h_i$ represents $h_i$ is selected as a better hypothesis.
\begin{equation}
    \fontsize{9.5pt}{11pt}\selectfont
    H = \{ h_1, h_2, \ldots, h_n \mid R(h_i, h_{i+1})=h_i \text{ for all } i \}
\label{eq:ranking_pairwise}
\end{equation}

\subsection{LLM-Based Agentic Framework}
\label{subsec:agentic_framework}
\begin{figure*}[t]
\centering
\resizebox{2\columnwidth}{!}{
\includegraphics[]{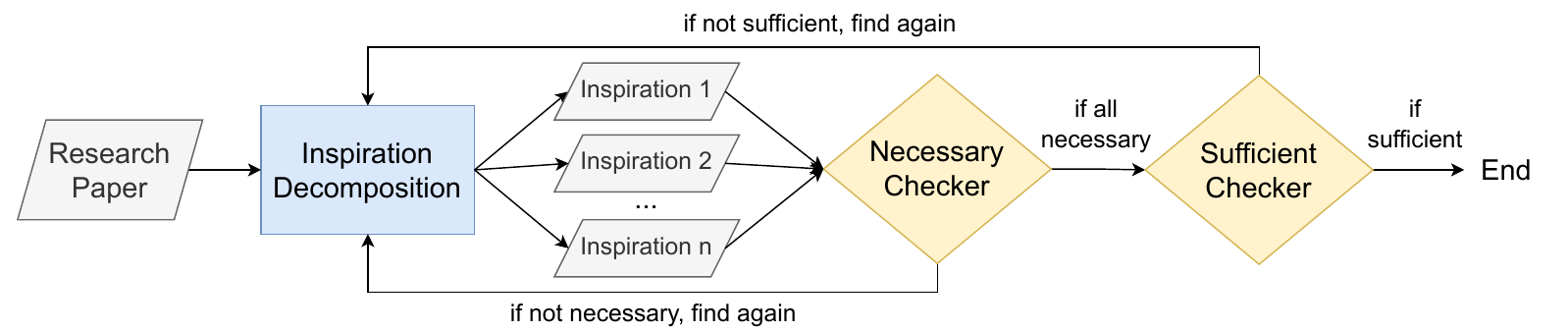}
}
\caption{Overview of the inspiration retrieval framework.}
\label{fig:framework}
\end{figure*}

We develop a LLM-based agentic framework to automatedly extract the research question, background survey, inspirations, and hypothesis. 

The extraction of research question, background survey, and hypothesis is relatively straightforward for LLMs.
Specifically, we carefully design prompts and adopt iterative self-refine~\citep{selfrefine} to extract them.
The background survey is summarized based on the information in the introduction section and the related work section.

The extraction of inspirations is not so straightforward compared to the other components. 
Figure~\ref{fig:framework} shows the inspiration extraction framework. 
We have simplified its design as much as possible, retaining only the essential components to ensure efficiency and accuracy.
To assist readers in understanding the practical input and output of this framework, we provide a concrete example in \ref{sec:example_pipeline}.

The inspirations are mostly described in the introduction section, usually starting with ``Motivated by'', and can be also described in the related work and methodology sections.
As shown in Figure~\ref{fig:framework}, given the full passage of a research paper, the ``inspiration decomposition'' module first iteratively extracts several potential inspirations. 
Here each potential inspiration is represented by the title and abstract of a referenced research paper.
Therefore after the ``inspiration decomposition'' module suggests the title of a referred paper as inspiration, we use Semantic Scholar and Crossref to find the abstract of the referred paper to compose an inspiration.
Then the ``necessary checker'' module examines whether each extracted inspiration is needed to formulate the hypothesis and not redundant, and the ``sufficient checker'' is to check whether all necessary inspirations have been extracted to be enough to be possible to formulate the hypothesis.
Here by ``enough'', we mean the information in the research question, background survey, and the inspirations can cover the information scope of the hypothesis.

To prevent data contamination, we apply the agentic framework only to papers published in 2024 or later, thereby minimizing overlap with the pretraining data of LLMs.
Furthermore, we provide a detailed data contamination analysis in Appendix \ref{sec:contamination_analysis}, where we verify model performance on a stricter subset that completely avoids data contamination.

\subsection{Negative Inspiration Selection}
\label{subsec:negative_insp_select}

Although ground-truth inspirations can be extracted, we need negative inspirations to calculate the performance of LLMs on inspiration retrieval. 
Here our goal is to provide an in-depth analysis on the inspiration retrieval ability by carefully composing a negative inspiration paper set for each paper in the benchmark.

Specifically, for each paper in the benchmark, we collect three levels of negative inspiration papers, based on their distance to the benchmark paper.
The first-level is papers that are cited and referred to by the benchmark paper, or papers that have high semantic similar titles to the benchmark paper. 
For each benchmark paper, we collect 100 citation-adjacent papers with Crossref API, and 50 semantic adjacent papers with Semantic Scholar API. 

The second-level are papers that are in the same discipline with the benchmark paper, and the third-level are papers that belong to completely different disciplines~(randomly selected).
\begin{table*}[ht]
    \centering
    
    \begin{subtable}{\textwidth}
        \centering
        \caption{The accuracy~(\%) of LLMs in retrieving the groundtruth inspiration while only \textbf{20\%} of inspiration candidates are selected.}
        \resizebox{\textwidth}{!}{
        \begin{tabular}{lccccccccccccc}
    \toprule
    \textbf{Model} &     \textbf{Cell} &        \textbf{Chem} &    \textbf{ETS} & \textbf{MS} &          \textbf{Phys} &   \textbf{EGS} & \textbf{EVS} &          \textbf{BL} &         \textbf{BS} &              \textbf{Law} &             \textbf{Math} &        \textbf{A} &      \textbf{Overall} \\
    \midrule
    Llama-3.2-1B & 34.65 & 34.80 & 32.57 & 30.26 & 30.25 & 34.75 & 35.43 & 33.21 & 41.09 & 29.74 & 36.22 & 30.10 & 33.68 \\
    Llama-3.1-8B & 74.08 & 78.00 & 79.69 & 74.54 & 76.75 & 84.56 & 75.20 & 75.81 & 80.00 & 65.95 & 75.59 & 68.37 & 75.92 \\
    Qwen Turbo & 74.37 & 77.20 & 80.08 & 72.69 & 75.80 & 88.03 & 78.35 & 74.01 & 82.18 & 67.24 & 74.80 & 66.84 & 76.17 \\
    GPT-4o Mini & 76.06 & 83.20 & 82.76 & 77.49 & 81.53 & 89.96 & 79.92 & 70.76 & 84.00 & 70.69 & 74.80 & 71.94 & 78.74 \\
    Gemini 2.0 FT & 74.65 & 79.60 & 80.84 & 73.43 & 78.34 & 90.35 & 76.77 & 75.09 & 85.09 & \textbf{80.17} & 76.38 & \textbf{77.55} & 78.89 \\
    Gemini 2.0 Flash & 75.77 & 76.40 & 85.82 & 75.28 & 79.94 & 91.89 & 75.98 & 75.09 & 86.91 & 78.02 & 76.77 & 71.94 & 79.24 \\
    Qwen Plus & 79.15 & 82.00 & 82.76 & 75.28 & 80.57 & 91.12 & 81.10 & 76.53 & 84.73 & 75.00 & 79.53 & 73.98 & 80.27 \\
    DeepSeek-V3 & 80.00 & 83.60 & 85.44 & 76.01 & 79.94 & 91.51 & 79.53 & 76.90 & 86.91 & 75.86 & 77.56 & 73.98 & 80.74 \\
    Claude 3.5 Haiku & \textbf{80.56} & 85.20 & 85.06 & 77.86 & 79.94 & 90.35 & 83.07 & 75.81 & 87.27 & 70.69 & 77.56 & 75.51 & 80.89 \\
    Llama-3.1-70B & 78.31 & 84.00 & 84.67 & 80.07 & 80.25 & 89.58 & 81.10 & \textbf{79.42} & 86.91 & 75.43 & 77.95 & 75.51 & 81.18 \\
    Claude 3.5 Sonnet & 78.31 & 78.40 & 85.06 & 76.75 & 81.53 & 91.51 & \textbf{85.04} & 77.62 & \textbf{88.00} & 77.59 & 79.53 & \textbf{77.55} & 81.43 \\
    GPT-4o & 80.00 & \textbf{87.20} & \textbf{89.27} & \textbf{80.81} & \textbf{84.39} & \textbf{93.05} & 81.89 & 77.98 & 87.64 & 79.74 & \textbf{83.07} & 75.00 & \textbf{83.43} \\
    \bottomrule
  \end{tabular}}
    \end{subtable}
    \vspace{1em} 
    
    \begin{subtable}{\textwidth}
        \centering
        \caption{The accuracy~(\%) of LLMs in retrieving the groundtruth inspiration while only \textbf{4\%} of inspiration candidates are selected.}
        \resizebox{\textwidth}{!}{
            \begin{tabular}{lccccccccccccc}
        \toprule
        \textbf{Model} &     \textbf{Cell} &        \textbf{Chem} &    \textbf{ETS} & \textbf{MS} &          \textbf{Phys} &   \textbf{EGS} & \textbf{EVS} &          \textbf{BL} &         \textbf{BS} &              \textbf{Law} &             \textbf{Math} &        \textbf{A} &      \textbf{Overall} \\
        \midrule
        Llama-3.2-1B & 10.70 & 11.60 & 12.26 & 9.59 & 11.15 & 8.49 & 14.57 & 13.00 & 17.09 & 12.50 & 11.42 & 10.71 & 11.91 \\
        Llama-3.1-8B & 32.39 & 38.00 & 40.61 & 31.37 & 32.80 & 59.85 & 36.61 & 28.52 & 55.64 & 28.88 & 36.22 & 34.69 & 37.87 \\
        Gemini 2.0 FT & 31.27 & 41.20 & 40.61 & 30.63 & 32.48 & 71.04 & 39.37 & 33.57 & 59.64 & 37.07 & 34.65 & 33.16 & 40.18 \\
        GPT-4o Mini & 30.42 & 43.60 & 41.00 & 34.69 & 33.44 & 66.80 & 40.16 & 28.88 & 64.73 & 32.76 & 37.80 & 35.71 & 40.59 \\
        Qwen Turbo & 35.49 & 42.40 & 42.15 & 33.95 & 35.03 & 66.80 & 43.31 & 33.21 & 61.45 & 29.74 & 36.61 & 34.69 & 41.21 \\
        Gemini 2.0 Flash & 31.55 & 38.80 & 44.06 & 34.32 & 34.39 & 74.52 & 37.40 & 32.49 & 64.00 & 37.50 & 37.80 & 32.65 & 41.46 \\
        Claude 3.5 Sonnet & 36.34 & 41.20 & 42.91 & 30.63 & 36.31 & 67.57 & 40.55 & 34.30 & 63.64 & 34.91 & 37.40 & 33.67 & 41.62 \\
        Qwen Plus & 36.06 & 47.20 & 45.21 & 33.58 & 34.39 & 72.97 & 43.31 & 35.38 & 64.36 & 34.91 & 39.37 & 36.22 & 43.43 \\
        Claude 3.5 Haiku & 41.13 & \textbf{48.40} & 45.98 & 34.69 & 33.44 & 69.88 & 44.09 & 34.30 & 64.00 & 37.93 & 38.19 & \textbf{41.33} & 44.28 \\
        DeepSeek-V3 & 38.87 & 46.00 & 44.06 & 36.90 & \textbf{36.62} & \textbf{75.29} & 41.73 & \textbf{40.07} & 65.45 & 36.64 & 38.58 & 37.76 & 44.78 \\
        Llama-3.1-70B & \textbf{41.41} & 44.00 & \textbf{47.51} & 36.90 & 34.39 & 70.66 & \textbf{45.28} & 37.18 & 65.45 & \textbf{39.22} & 38.19 & 39.29 & 44.87 \\
        GPT-4o & 39.44 & 46.40 & 47.13 & \textbf{38.38} & 35.35 & \textbf{75.29} & 44.88 & 38.63 & \textbf{65.82} & \textbf{39.22} & \textbf{40.16} & 38.78 & \textbf{45.65} \\
        \bottomrule
      \end{tabular}}
    \end{subtable}
    
    \caption{Performance of LLMs in hypothesis retrieve task. Gemini 2.0 FT=Gemini 2.0 Flash Thinking; Chem=Chemistry, ETS=Earth Science, MS=Material Science, Phys=Physics, EGS=Energy Science, EVS=Environmental Science, BL=Biology, BS=Business, A=Astronomy.}
    \label{tab:retrieve_r1_r2}
\end{table*}

We randomly collect 2000 papers for each discipline by Web of Science, which can be used for both the second-level and the third-level. 
During experiment, for each benchmark paper, we randomly select 25 negative inspiration papers from each of the distance level to compose the negative inspiration set.

The purpose of the three-level design is two-fold.
Firstly, real inspiration can come from each of the levels. 
By the splitting, LLM's preference in terms of distance can be analyzed.
Secondly, the incorporation of closely related papers makes the negative inspiration papers non-trivial: we find that LLMs tend to select papers that are close to the benchmark paper.
If the negative papers are only from irrelevant disciplines, then the retrieval success rate will be very high and less meaningful.

\begin{table*}[h]
  \centering
  \resizebox{1.60\columnwidth}{!}{
  \begin{tabular}{l@{\hspace{5pt}}c@{\hspace{10pt}}c@{\hspace{10pt}}c@{\hspace{10pt}}c@{\hspace{10pt}}c@{\hspace{10pt}}c}
    \toprule
    \textbf{Model} & \multicolumn{2}{l}{\textbf{Distance Level 1}} & \multicolumn{2}{l}{\textbf{Distance Level 2}} & \multicolumn{2}{l}{\textbf{Distance Level 3}} \\
                   & \textbf{(top 20\%)} & \textbf{(top 4\%)} & \textbf{(top 20\%)} & \textbf{(top 4\%)} & \textbf{(top 20\%)} & \textbf{(top 4\%)} \\
    \midrule
    Llama-3.2-1B & 23.57\% & 6.33\% & 15.52\% & 2.93\% & 14.46\% & 2.85\% \\
    Qwen Turbo & 52.72\% & 12.05\% & 9.45\% & 1.11\% & 4.46\% & 0.34\% \\
    Claude 3.5 Sonnet & 53.96\% & 10.15\% & 10.16\% & 0.70\% & 2.40\% & 0.13\% \\
    Llama-3.1-8B & 53.69\% & 11.17\% & 10.65\% & 0.77\% & 2.94\% & 0.14\% \\
    Gemini 2.0 Flash Thinking & 54.49\% & 10.59\% & 10.34\% & 0.58\% & 2.24\% & 0.11\% \\
    GPT-4o & 54.90\% & 10.02\% & 9.84\% & 0.47\% & 2.09\% & 0.09\% \\
    Llama-3.1-70B & 55.32\% & 10.04\% & 9.82\% & 0.55\% & 2.16\% & 0.09\% \\
    DeepSeek-V3 & 55.74\% & 10.22\% & 9.80\% & 0.43\% & 1.79\% & 0.07\% \\
    GPT-4o Mini & 55.90\% & 10.67\% & 9.54\% & 0.47\% & 2.12\% & 0.09\% \\
    Claude 3.5 Haiku & 55.70\% & 10.19\% & 9.51\% & 0.49\% & 2.00\% & 0.07\% \\
    Gemini 2.0 Flash & 55.91\% & 10.63\% & 9.63\% & 0.42\% & 2.03\% & 0.09\% \\
    Qwen Plus & 56.11\% & 10.57\% & 9.52\% & 0.50\% & 2.16\% & 0.16\% \\
    \bottomrule
  \end{tabular}}
  \caption{Analysis of negative inspiration retrieval in the inspiration retrieval task. Each value represents the average percentage of negative inspirations retrieved across three distance levels, under two settings where only 20\% and 4\% of the candidate inspirations are selected, respectively.}
  \label{tab:retrieve_negative_inspiration}
\end{table*}

\subsection{Expert Evaluation}
\label{subsec:expert_evaluation}

We invited five PhD students from diverse disciplines—Physics (1), Chemistry (2), Materials Science (1), and Astronomy (1)—to evaluate the accuracy of our decomposition framework. Specifically, we randomly sampled 62 papers from the benchmark dataset, each accompanied by its extracted research question, background survey, inspirations, and hypothesis, and presented them in the form of a questionnaire.
Detailed information about annotator recruitment, the annotation procedure, and the specific guidelines are provided in Appendix \ref{appendix_prompt_for_expert_checking}.

For each paper, the experts first read the full text of the original research paper and then assessed the accuracy of the extracted components. Each inspiration was evaluated for its necessity, while the entire set of inspirations was assessed for its sufficiency in supporting the research hypothesis.

Overall, the evaluation identified five cases with issues in inspiration identification~(three) or hypothesis extraction~(two), and six minor issues in research question extraction. The decomposition accuracy was 91.9\% considering only major issues and 82.3\% when including all issues.








\section{Experiments}

We evaluate twelve representative LLMs in our experiments: Llama-3.2-1B~\citep{llama3}, Llama-3.1-8B, Llama-3.1-70B, Gemini 2.0 Flash~\citep{gemini20}, Gemini 2.0 Flash Thinking,  Qwen Turbo~\citep{qwen2}, Qwen Plus, Claude 3.5 Haiku~\citep{claude3}, Claude 3.5 Sonnet,  DeepSeek-V3~\citep{deepseekv3}, GPT-4o Mini~\cite{gpt4o} and GPT-4o.

For a better understanding of these tasks, we provide concrete input and output examples for each of the three subtasks in Appendix \ref{sec:subtask_examples}.


\begin{figure*}[t]
\centering
\resizebox{1.8\columnwidth}{!}{
\includegraphics[]{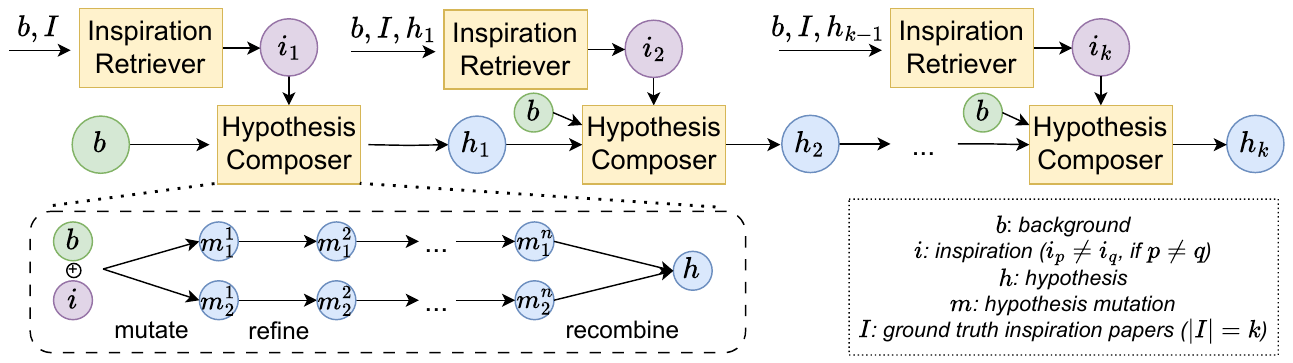}
}
\caption{Overview of the hypothesis composition process.}
\label{fig:hyp_gene}
\end{figure*}

\begin{table*}[h]
  \centering
  \resizebox{1.85\columnwidth}{!}{
  \begin{tabular}{lccccccccccccc}
    \toprule
        \textbf{Model} &     \textbf{Cell} &        \textbf{Chem} &    \textbf{ETS} & \textbf{MS} &          \textbf{Phys} &   \textbf{EGS} & \textbf{EVS} &          \textbf{BL} &         \textbf{BS} &              \textbf{Law} &             \textbf{Math} &        \textbf{A} &      \textbf{Overall} \\
    \midrule
    Claude 3.5 Haiku & 40.42 & 40.87 & 38.71 & 46.75 & 45.00 & 45.34 & 48.00 & 46.15 & 35.14 & 37.85 & 43.59 & 34.29 & 42.56 \\
    Llama-3.1-8B & 44.58 & 47.83 & 42.78 & 46.04 & 45.05 & 44.30 & 46.47 & 47.37 & 44.21 & 47.58 & 48.21 & 45.14 & 45.68 \\
    Gemini 2.0 FT & 45.67 & 39.79 & 48.48 & 47.22 & 48.77 & 49.24 & 48.57 & 48.02 & 41.47 & 47.03 & 42.81 & 40.00 & 46.30 \\
    Gemini 2.0 Flash & 46.25 & 45.63 & 48.64 & 51.63 & 47.97 & 51.47 & 49.41 & 48.77 & 47.03 & 55.91 & \textbf{56.24} & 49.71 & 50.15 \\
    Llama-3.1-70B & 46.67 & 49.86 & 50.83 & 51.53 & 50.60 & 50.61 & 52.10 & 54.36 & 49.47 & 53.94 & 51.11 & 49.14 & 50.92 \\
    GPT-4o Mini & 46.67 & 49.42 & 50.91 & 52.63 & 53.82 & 53.33 & 54.86 & 54.36 & 46.92 & 56.97 & 52.48 & 53.14 & 52.47 \\
    Qwen Turbo & 52.92 & 51.45 & 49.55 & 51.06 & 52.64 & 50.97 & 52.57 & 56.92 & \textbf{53.16} & 55.76 & 55.38 & 53.14 & 52.71 \\
    GPT-4o & 55.00 & 53.04 & 54.09 & 53.95 & 53.82 & 52.97 & 53.14 & 55.38 & 46.15 & 53.99 & 54.53 & 52.57 & 53.37 \\
    DeepSeek-V3 & 52.78 & 52.27 & 53.18 & 54.25 & 54.91 & 53.91 & 53.71 & 56.32 & 50.27 & 55.15 & 52.14 & 53.71 & 53.79 \\
    Qwen Plus & \textbf{60.00} & \textbf{53.72} & \textbf{57.27} & \textbf{56.63} & \textbf{58.14} & \textbf{56.63} & \textbf{60.57} & \textbf{58.97} & 51.05 & \textbf{62.19} & 55.90 & \textbf{56.57} & \textbf{57.46} \\
    \bottomrule
  \end{tabular}}
  \caption{Performance of LLMs in hypothesis composition task. Each number represents the normalized performance of LLMs in composing hypothesis. Gemini 2.0 FT=Gemini 2.0 Flash Thinking; Chem=Chemistry, ETS=Earth Science, MS=Material Science, Phys=Physics, EGS=Energy Science, EVS=Environmental Science, BL=Biology, BS=Business, A=Astronomy.}
  \label{tab:generation_score}
\end{table*}

\subsection{Inspiration Retrieval}
\label{inspiration_retrieval}

For each benchmark paper, we use the extracted research question, groundtruth inspirations, and negative inspirations to evaluate LLMs’ retrieval accuracy. Each candidate set contains 75 papers (2–3 groundtruth and 25 negatives per distance level), represented by title and abstract. 
Retrieval proceeds in two rounds. In Round 1, the 75 candidates are partitioned into five groups of 15; within each group, the LLM selects the top~3, yielding 15(\(5\times3\)) papers for Round 2. In Round 2, these 15 papers are evaluated together (one group) and the LLM selects the top~3 overall, leaving 4\%(\(3/75\)) of the original set.
We use Hit Ratio—the fraction of groundtruth inspirations among selected ones—as the evaluation metric. 
The Hit Ratio results are presented in Table~\ref{tab:retrieve_r1_r2}. 
The values in the ``Overall'' column represent averages across 12 disciplines and 1,386 benchmark papers. Overall, LLMs demonstrate surprisingly high retrieval accuracy.

In the first round (20\% retained), most models identify about 80\% of groundtruth inspirations; even in the final round (4\% retained), where only 3 papers are selected, accuracy remains above 40\%, with GPT-4o leading at 45.65\%. These findings show that LLMs can identify papers not explicitly related to the research question but potentially valuable for hypothesis formation, likely because their pretraining process captured latent cross-domain associations not recognized by scientists.



Table~\ref{tab:retrieve_r1_r2} also indicate the scaling law of LLMs for inspiration retrieval: the inspiration retrieval ability grows up very fast before and during 8B parameters, while stuck in a bottleneck at around 70B parameters. 
No matter how the LLMs are trained with different strategies, they seem to be bottlenecked at the same performance.

We further analyze the Hit Ratio of negative inspirations across three distance levels (Table~\ref{tab:retrieve_negative_inspiration}).
Results show that closer papers are more likely to be selected, regardless of selection percentage.
This trend arises because nearby papers are statistically more relevant and frequently co-occur in pretraining corpora, biasing LLMs toward selecting them as inspirations. 

\begin{table*}[h]
  \centering
  \resizebox{1.85\columnwidth}{!}{
  \begin{tabular}{lccccccccccccc}
\toprule
                    \textbf{Model} &     \textbf{Cell} &        \textbf{Chem} &    \textbf{ETS} & \textbf{MS} &          \textbf{Phys} &   \textbf{EGS} & \textbf{EVS} &          \textbf{BL} &         \textbf{BS} &              \textbf{Law} &             \textbf{Math} &        \textbf{A} &      \textbf{Overall} \\
\midrule
Llama-3.1-70B & 36.94 & 35.57 & 30.57 & 37.71 & 43.35 & 47.18 & 36.02 & 43.11 & 41.63 & 46.09 & 30.73 & 25.40 & 38.06 \\
GPT-4o Mini & 42.25 & 39.94 & 34.39 & 42.98 & 39.78 & 43.78 & 40.63 & 43.72 & 45.03 & 42.24 & 32.67 & 31.50 & 40.13 \\
Gemini 2.0 Flash & 43.73 & 44.38 & 35.95 & 51.86 & 54.63 & 55.16 & 40.98 & 44.00 & 46.88 & 48.31 & 38.24 & 35.75 & 45.11 \\
Qwen Turbo & 46.42 & 45.11 & 42.88 & 48.74 & 45.61 & 46.40 & 45.26 & 49.20 & 50.92 & 49.27 & 37.15 & 37.62 & 45.48 \\
Gemini 2.0 FT & 43.52 & 44.96 & 36.88 & 52.81 & 54.08 & 54.95 & 42.27 & 44.53 & 46.15 & 48.09 & 37.80 & 38.40 & 45.49 \\
Qwen Plus & 46.00 & 46.00 & 41.72 & 49.35 & 50.64 & 49.11 & 44.80 & 46.93 & 43.36 & 45.43 & 40.16 & 41.97 & 45.56 \\
Claude 3.5 Haiku & 48.15 & 46.88 & 45.55 & 52.45 & 54.10 & 52.48 & 48.83 & 48.06 & 51.23 & 52.93 & 44.49 & 40.27 & 48.86 \\
Llama-3.1-8B & 55.48 & 54.20 & 55.90 & 56.60 & 54.35 & 55.48 & 55.91 & 56.71 & 54.69 & 55.55 & 55.60 & 55.49 & 55.65 \\
GPT-4o & 60.75 & 60.99 & 53.24 & 61.69 & 61.34 & 61.20 & 60.52 & 64.11 & 64.67 & 61.14 & 52.60 & 51.80 & 59.60 \\
DeepSeek-V3 & \textbf{80.88} & \textbf{82.03} & 78.85 & \textbf{83.63} & 80.82 & 81.47 & \textbf{83.98} & 81.77 & 83.48 & 80.69 & \textbf{76.78} & 75.88 & 80.99 \\
Claude 3.5 Sonnet & 80.23 & 80.83 & \textbf{80.93} & 83.20 & \textbf{84.33} & \textbf{84.72} & 82.63 & \textbf{82.48} & \textbf{84.87} & \textbf{81.81} & 76.20 & \textbf{76.51} & \textbf{81.59} \\
\bottomrule
\end{tabular}}
  \caption{Performance of LLMs in hypothesis ranking task. Each number represents the accuracy~(\%) of LLMs in ranking ground-truth hypothesis among negative hypothesis. Chem=Chemistry, ETS=Earth Science, MS=Material Science, Phys=Physics, EGS=Energy Science, EVS=Environmental Science, BL=Biology, BS=Business, A=Astronomy.}
  \label{tab:ranking_accuracy}
\end{table*}

\subsection{Hypothesis Composition}
\label{hypothesis_composition}

\begin{table}[h]
  \centering
  \small
  \setlength{\tabcolsep}{6pt}
  \begin{tabular}{lcccc}
    \toprule
    \textbf{Model} & \textbf{\ding{55} \ding{55}} & \textbf{\ding{51} \ding{55}} & \textbf{\ding{51} \ding{51}}\\
    \midrule
GPT-4o Mini & 33.83 & 64.83 & 1.33\\
Qwen Plus & 25.00 & 69.33 & 5.67\\
Llama-3.1-8B & 2.50 & \textbf{91.67} & 5.83\\
Llama-3.1-70B & \textbf{52.67} & 39.17 & 8.17\\
Gemini 2.0 Flash & 35.50 & 51.67 & 12.83 \\
Claude 3.5 Haiku & 28.17 & 58.17 & 13.67\\
Gemini 2.0 FT & 36.50 & 49.67 & 13.83\\
Qwen Turbo & 39.33 & 45.67 & 15.00 \\
GPT-4o & 11.50 & 61.50 & 27.00 \\
DeepSeek-V3 & 1.74 & 21.83 & 76.44\\
Claude 3.5 Sonnet & 3.17 & 19.17 & \textbf{77.67} \\
\bottomrule
  \end{tabular}
  \caption{Analysis of \textit{position bias} in hypothesis ranking task. Specifically, each hypothesis pair is compared twice, with three possible outcomes: both wrongly ranked~(\ding{55} \ding{55}); one right one wrong~(\ding{51} \ding{55}); both rightly ranked~(\ding{51} \ding{51}). Each number represents an averaged percentage~(\%).}
  \label{tab:hypothesis_ranking_position}
\end{table}


With the retrieved inspirations, the next step is to associate them with the research background to compose research hypothesis.
Figure \ref{fig:hyp_gene} shows the framework we use for the hypothesis generation process. 
This framework strictly follows Equation~\ref{eq:decomposition}, and is designed to be as simple as possible, avoiding unnecessary components.

We use the evolutionary unit~\citep{msc} to associate the research background~($b$) and inspiration~($i$), shown in the bottom-left rectangle in Figure~\ref{fig:hyp_gene}.
Specifically, ``mutate'' means creating different ways to combine $b$ and $i$ together and ``recombine'' tries to keep the merits of different combination ways to compose a final hypothesis.
The prompts for mutate, refine, and recombine are provided in Appendix \ref{Prompts for mutate, refine, and recombine}. In this step, we only measure the LLM's ability on hypothesis composing. 
For the LLM-generated hypotheses to be comparable with the groundtruth hypothesis for evaluation, here $I$ represents the groundtruth inspiration papers~(usually 2 to 3 papers), while ``inspiration retriever'' module each time retrieve only one inspiration, and will not retrieve the same inspiration again.

The detailed scoring criteria is shown in Appendix \ref{appendix_prompt_for_generation_Criteria}, where we use a 6-point Likert scale~(from 0 to 5) to measure whether the generated hypothesis has covered the key points in the groundtruth hypothesis.
To compute the generation accuracy, we normalize the average score by dividing it by the maximum possible score (5). The final results are summarized in Table \ref{tab:generation_score}. 
Table~\ref{tab:generation_score} shows that 
(a) all LLMs preserve a certain kind of ability to associate the research background and inspirations to compose hypothesis;
(b) the hypothesis composition task remains challenging, as none of the models achieve consistently high performance.

\subsection{Hypothesis Ranking}

In this section, we evaluate the ability of LLMs to rank hypotheses pairwisely. Specifically, based on the hypothesis generation method in \S~\ref{hypothesis_composition}, we use the top-ranked negative inspirations and background question to construct a set of negative hypotheses. From this set, we randomly sample 5 negative hypotheses. 
Additionally, with subsets of groundtruth inspirations and the research question, we use the hypothesis composition framework to generate another set of negative hypotheses. 
In this set, we randomly sample 10 as negative hypotheses for ranking. 
As a result, for each benchmark paper, we compose a set of 16 hypotheses, including one groundtruth one and 15 negative ones for ranking.
To evaluate ranking performance, we use the groundtruth hypothesis to pairwisely compare with each of the 15 negative ones. 
The prompt for pairwise ranking is provided in the Appendix~\ref{appendix_prompt_for_ranking}.

Accuracy is used as the evaluation metric, which is calculated as the proportion of correct pairwise rankings out of 15 comparisons. 
During the pairwise evaluation, we find that many LLMs have strong position bias: they largely prefer the first hypothesis than the second. 
To avoid this bias, for each hypothesis pair, we compare them twice with reverse positions, and the results are averaged.

Table~\ref{tab:ranking_accuracy} presents the ranking accuracy of each LLM.
This ranking results indicate a different scaling law with the scaling law we find in the inspiration retrieval task: more parameters and better pretraining strategies can significantly improve over the hypothesis ranking task, while might lead to less improvements in the inspiration retrieval task.

Table~\ref{tab:hypothesis_ranking_position} analyzes position bias in the hypothesis ranking task: each pair is compared twice (yielding three possible outcomes), with the table reporting averaged percentages per outcome. It shows many LLMs are heavily influenced (e.g., Llama-3.1-8B reaches self-contradictory results 91.67\% of the time), while some are less so (e.g., Claude 3.5 Sonnet at only 19.17\%). The high rate of self-contradictory results likely explains why many LLMs achieve $\sim$50\% accuracy in Table~\ref{tab:ranking_accuracy}.




\section{Analysis}

\subsection{LLMs as Research Hypothesis Mines}
Our results show that (1) LLMs capture latent knowledge associations, enabling accurate inspiration retrieval; (2) given correct inspirations, they compose hypotheses reflecting key elements of the original innovations; and (3) with larger scale and better training, their hypothesis ranking improves without clear limit. Given that these three tasks—inspiration retrieval, hypothesis composition, and ranking—form a sufficient set for scientific discovery, LLMs can autonomously discover hypotheses from only a research background: screening inspiration candidates to select the best, associating them with the background, and ranking the resulting hypotheses to deliver the top ones to scientists. In this view, LLMs function as research hypothesis mines: stronger models represent richer mines, while greater inference compute corresponds to deploying more miners.




\subsection{Error Analysis}
To analyze failure cases, we randomly selected 100 incorrect cases for each of the three sub-tasks and categorized their causes.
For inspiration retrieval, 56\% of errors were because the model judged relevance only by title or abstract overlap, ignoring the conceptual link to the research question.
23\% were because the model missed papers from other disciplines that used different terms but had helpful inspiration.
4\% resulted from misunderstanding the research question itself, leading to mismatched selections.
The remaining 17\% were due to other minor factors.
For hypothesis composition, most errors fall into three categories:
52\% were due to missing key elements, where the model ignored important information from the background or inspirations, leading to incomplete hypotheses;
27\% resulted from poor combination of inspirations, where the model combines the information only superficially without building a clear cause;
13\% occurred when the generated hypothesis was fluent but misaligned with the original research question; the remaining 8\% were minor issues.

For hypothesis ranking, 83\% of errors arise from how models process the order of comparison rather than the content itself.
Another 11\% occur when both hypotheses are reasonable, and the model struggles to make subtle distinctions in how well each fits the research question.
The remaining 6\% are due to other minor factors.

\subsection{Insights \& Challenges}
In inspiration retrieval, the limitation stems from pretraining corpora, which prioritize domain coverage over cross-domain connectivity, yielding dense intra-domain co-occurrences but sparse inter-domain links and thus restricting retrieval of weakly related yet potentially groundbreaking inspirations; the bottleneck lies in training data distribution, not model reasoning. 

In hypothesis composition, the limitation arises from the optimization objective of maximizing \(P(x_t | x_{<t})\), which favors outputs following frequent training patterns, while novel hypotheses—weakly represented concept combinations—have inherently low likelihood under the learned distribution, leading models to prefer semantically coherent but familiar compositions over exploratory or unconventional ones. 

In hypothesis ranking, performance is constrained by the autoregressive architecture, where predictions depend on preceding tokens, creating positional dependencies that bias models toward the first input in comparisons and yield systematic bias in pairwise ranking.







\section{Conclusion}

We present ResearchBench, a large-scale benchmark for LLM-driven scientific discovery across inspiration retrieval, hypothesis composition, and hypothesis ranking, spanning 12 disciplines and built with an automated LLM framework for scalable, contamination-resistant construction. LLMs show strong retrieval but only moderate composition and ranking, indicating they capture useful associations yet lack deeper integrative reasoning. Addressing these bottlenecks could turn LLMs into “research hypothesis mines,” enabling autonomous generation of high-quality hypotheses and advancing automated scientific discovery.

\section*{Limitations}
Due to budget and resource constraints, ResearchBench currently covers only 12 disciplines. The included disciplines were selected primarily based on data and resource availability rather than deliberate curation, and no discipline was excluded post hoc because it appeared not to admit decomposition under the theoretical foundation described in \S~\ref{sec:preliminary}.

In addition, ResearchBench is not specifically designed for fine-grained hypothesis discovery, a task introduced and formally formulated in MOOSE-Chem2~\citep{msc2}. Extending it to this setting would require applying the automated agentic extraction framework to derive fine-grained research hypotheses from the literature as labels, together with evaluation metrics tailored to fine-grained hypothesis composition.

ResearchBench is also not designed to evaluate a discovery system's ability (e.g., that of an LLM) to leverage external experimental feedback for hypothesis updating~\citep{msc3,funsearch,alphaevolve,llmsr,deepscientist}.

Finally, while the automatically collected labels from the literature may support model training, this direction remains underexplored. Although \citet{mss} has enabled scalable and tractable training of $P(\text{hypothesis}\mid\text{background})$, substantial room remains for further research on training paradigms for scientific discovery.


\section*{Acknowledgments}
This work is supported by New Generation Artificial Intelligence-National Science and Technology Major Project(2025ZD0121802).

\bibliography{custom}

\newpage
\appendix
\onecolumn

\section{Appendix}
\label{sec:appendix}

\subsection{Data Contamination Analysis}
\label{sec:contamination_analysis}

To ensure the validity of our benchmark, we meticulously address the risk of data contamination. For closed-source models, we minimize this risk by selecting the earliest available versions for evaluation. 
The knowledge cutoff dates and release dates for all evaluated models are detailed in Table~\ref{tab:model_cutoff}.

\begin{table}[h]
\centering
\small
\begin{tabular}{l|c|c}
\toprule
\textbf{Model} & \textbf{Cutoff Date} & \textbf{Release Date} \\
\midrule
GPT-4o & Oct 2023 & May 2024 \\
GPT-4o Mini & Oct 2023 & Jul 2024 \\
Llama-3.1-8B & Dec 2023 & Jul 2024 \\
Llama-3.1-70B & Dec 2023 & Jul 2024 \\
Gemini 2.0 Flash & Jun 2024 & Dec 2024 \\
Gemini 2.0 FT & Jun 2024 & Dec 2024 \\
Claude 3.5 Sonnet & Apr 2024 & Jun 2024 \\
Claude 3.5 Haiku & Jul 2024 & Oct 2024 \\
Qwen Plus & - & Nov 2024 \\
Qwen Turbo & - & Nov 2024 \\
DeepSeek-V3 & - & Dec 2024 \\
\bottomrule
\end{tabular}
  \caption{LLM's pretraining data cutoff date. Symbol `-' means the official cutoff date is not specified.}
  \label{tab:model_cutoff}
\end{table}


In addition, to further verify that our benchmark is not affected by potential contamination near model cutoff dates, we rerun all three tasks on a stricter subset. This subset includes only papers published after July 2024, ensuring they are farther away from the baseline model cutoff dates. 

In the tables below (Table \ref{tab:strict_ir}, \ref{tab:strict_hc}, and \ref{tab:strict_rank}), \textbf{Original} refers to results on the full 2024 benchmark, while \textbf{Strict} refers to results on the post-July 2024 subset. Models marked with $\dag$ have a cutoff date strictly before July 2024, meaning the strict subset is theoretically invisible to them during pre-training.

\begin{table}[h]
\centering
\small
\begin{tabular}{l|cc}
\toprule
\textbf{Model} & \textbf{Original} & \textbf{Strict} \\
\midrule
Llama-3.2-1B$^\dag$ & 11.91 & 11.44 \\
Llama-3.1-8B$^\dag$ & 37.87 & 37.57 \\
Gemini 2.0 FT$^\dag$ & 40.18 & 39.88 \\
GPT-4o Mini$^\dag$ & 40.59 & 40.02 \\
Gemini 2.0 Flash$^\dag$ & 41.46 & 42.41 \\
Claude 3.5 Sonnet$^\dag$ & 41.62 & 42.41 \\
Claude 3.5 Haiku$^\dag$ & 44.28 & 43.67 \\
Llama-3.1-70B$^\dag$ & 44.87 & 44.30 \\
GPT-4o$^\dag$ & 45.65 & 45.22 \\
Qwen Turbo & 41.21 & 40.58 \\
Qwen Plus & 43.43 & 42.97 \\
DeepSeek-V3 & 44.78 & 43.83 \\
\bottomrule
\end{tabular}
\caption{Comparision on Inspiration Retrieval ($\dag$ = cutoff strictly before Jul 2024)}
\label{tab:strict_ir}
\end{table}

\begin{table}[h]
\centering
\small
\begin{tabular}{l|cc}
\toprule
\textbf{Model} & \textbf{Original} & \textbf{Strict} \\
\midrule
Claude 3.5 Haiku$^\dag$ & 42.56 & 41.13 \\
Llama-3.1-8B$^\dag$ & 45.68 & 44.76 \\
Gemini 2.0 FT$^\dag$ & 46.30 & 46.53 \\
Gemini 2.0 Flash$^\dag$ & 50.15 & 49.88 \\
Llama-3.1-70B$^\dag$ & 50.92 & 50.71 \\
GPT-4o Mini$^\dag$ & 52.47 & 51.59 \\
GPT-4o$^\dag$ & 53.37 & 53.16 \\
Qwen Turbo & 52.71 & 53.08 \\
DeepSeek-V3 & 53.79 & 53.36 \\
Qwen Plus & 57.46 & 56.24 \\
\bottomrule
\end{tabular}
\caption{Comparision on Hypothesis Composition ($\dag$ = cutoff strictly before Jul 2024)}
\label{tab:strict_hc}
\end{table}

\begin{table}[h]
\centering
\small
\begin{tabular}{l|cc}
\toprule
\textbf{Model} & \textbf{Original} & \textbf{Strict} \\
\midrule
GPT-4o Mini$^\dag$ & 1.33 & 2.60 \\
Llama-3.1-8B$^\dag$ & 5.83 & 10.15 \\
Llama-3.1-70B$^\dag$ & 8.17 & 13.84 \\
Gemini 2.0 Flash$^\dag$ & 12.83 & 20.42 \\
Claude 3.5 Haiku$^\dag$ & 13.67 & 21.47 \\
Gemini 2.0 FT$^\dag$ & 13.83 & 21.56 \\
GPT-4o$^\dag$ & 27.00 & 34.99 \\
Claude 3.5 Sonnet$^\dag$ & 77.67 & 79.05 \\
Qwen Plus & 5.67 & 4.57 \\
Qwen Turbo & 15.00 & 22.84 \\
DeepSeek-V3 & 76.44 & 72.36 \\
\bottomrule
\end{tabular}
\caption{Comparision on Hypothesis Ranking ($\dag$ = cutoff strictly before Jul 2024)}
\label{tab:strict_rank}
\end{table}

From the tables above, we can see that the results on the post-July 2024 subset are very close to those on the full 2024 benchmark. These minor differences are consistent with normal sample variance. This stability shows that our method effectively avoids data contamination.

\subsection{Expert Evaluation Details and Guidelines}

\label{appendix_prompt_for_expert_checking}

To ensure the high quality and correctness of the extracted components in our benchmark, we conducted a rigorous human evaluation. Below we detail the annotator recruitment, annotation procedure, and the specific guidelines used.

\textbf{Annotator Recruitment.} We invited five PhD students from Physics, Chemistry, Materials Science, and Astronomy, to serve as expert annotators. They check the accuracy of the extracted research questions, inspirations, and hypotheses, and assessed the necessity and sufficiency of the inspirations for recomposing the hypothesis.

\noindent\textbf{Annotation Procedure.} The validation process followed three steps. First, each expert selected papers from the benchmark that fell strictly within their own research expertise and familiarity. Then, experts were required to read the full text of the original research paper to understand the complete context. Finally, experts compared the content extracted by our framework (Research Question, Inspirations, Hypothesis) against the original paper content using the guidelines provided below.

Regarding inter-annotator agreement, it is difficult for an expert to accurately evaluate papers across different disciplines (e.g., an expert in Quantum Physics may not be qualified to evaluate a paper on Organic Chemistry). Therefore, we assigned papers based on the experts' specific research domains, so they rarely reviewed the same papers.

\textbf{Guideline.} The specific guideline used by experts to check the decomposition is provided below:

Titile:

Background question decomposed by automated framework:

Whether the background question correct?

<Reply fill in here. Required a detailed analysis>

ground-truth hypothesis decomposed by automated framework: 

Whether the ground-truth hypothesis accurately reflect the main proposal of the paper?

<Reply fill in here. Required a detailed analysis>

Inspiration paper 1 title:   Relation between the inspiration 1 paper and the decomposed paper:

Whether the collected inspiration paper 1 compose of a set of necessary conditions to reach to the ground-truth hypothesis?

<Reply fill in here. Required a detailed analysis>

Inspiration paper 2 title:   Relation between the inspiration 2 paper and the decomposed paper:

Whether the collected inspiration paper 2 compose of a set of necessary conditions to reach to the ground-truth hypothesis?

<Reply fill in here. Required a detailed analysis>

Inspiration paper 3 title:   Relation between the inspiration 3 paper and the decomposed paper:

Whether the collected inspiration paper 3 compose of a set of necessary conditions to reach to the ground-truth hypothesis?

<Reply fill in here. Required a detailed analysis>

Whether the collected inspirations paper compose of a set of sufficient conditions to reach to the coarse-grained hypothesis?

<Reply fill in here. Required a detailed analysis>

\subsection{Prompt for Retrieving Inspirations}
\label{appendix_prompt_retrieve}
You are helping with the scientific hypotheses generation process. Given a research question, the background and some of the existing methods for this research question, and several top-tier publications (including their title and abstract), try to identify which publication can potentially serve as an inspiration for the background research question so that combining the research question and the inspiration in some way, a novel, valid, and significant research hypothesis can be formed. The inspiration does not need to be similar to the research question. In fact, probably only those inspirations that are distinct with the background research question, combined with the background research question, can lead to a impactful research hypothesis. The reason is that if the inspiration and the background research question are semantically similar enough, they are probably the same, and the inspiration might not provide any additional information to the system, which might lead to a result very similar to a situation that no inspiratrions are found. An example is the backpropagation of neural networks. In backpropagation, the research question is how to use data to automatically improve the parameters of a multi-layer logistic regression, the inspiration is the chain rule in mathematics, and the research hypothesis is the backpropagation itself. In their paper, the authors have conducted experiments to verify their hypothesis. Now try to select inspirations based on background research question. The background research question is: ", "The introduction of the previous methods is:", "The potential inspiration candidates are: ", "Now you have seen the background research question, and many potential inspiration candidates. Please try to identify which three literature candidates are the most possible to serve as the inspiration to the background research question? Please name the title of the literature candidate, and also try to give your reasons.



















\subsection{Prompt for Evaluating Generated Hypothesis}
\label{appendix_prompt_for_generation_Criteria}
You are helping to evaluate the quality of a proposed research hypothesis by a phd student. The groundtruth hypothesis will also be provided to compare. Here we mainly focus on whether the proposed hypothesis has covered the key points of the ground-truth hypothesis. You will also be given a summary of the key points in the ground-truth hypothesis for reference. The evaluation criteria is called 'Matched score', which is in a 6-point Likert scale (from 5 to 0). Particularly, 5 points mean that the proposed hypothesis (1) covers three key points (or covers all the key points) in the ground-truth hypothesis, where every key point is leveraged nearly identically as in the ground-truth hypothesis, and (2) does not contain any extra key point(s) that is redundant, unnecessary, unhelpful, or harmful; 4 points mean that the proposed hypothesis (1) covers three key points (or covers all the key points) in the ground-truth hypothesis, where every key point is leveraged nearly identically as in the ground-truth hypothesis, and (2) but also contain extra key point(s) that is redundant, unnecessary, unhelpful, or harmful; 3 points mean that the proposed hypothesis (1) covers two key points in the ground-truth hypothesis, where every key point is leveraged nearly identically as in the ground-truth hypothesis, (2) but does not cover all key points in the ground-truth hypothesis, and (3) might or might not contain extra key points; 2 points mean that the proposed hypothesis (1) covers one key point in the ground-truth hypothesis, and leverage it nearly identically as in the ground-truth hypothesis, (2) but does not cover all key points in the ground-truth hypothesis, and (3) might or might not contain extra key points; 1 point means that the proposed hypothesis (1) covers at least one key point in the ground-truth hypothesis, but all the covered key point(s) are used differently as in the ground-truth hypothesis, and (2) might or might not contain extra key points; 0 point means that the proposed hypothesis does not cover any key point in the ground-truth hypothesis at all. Usually total the number of key points a ground-truth hypothesis contain is less than or equal to three. Please note that the total number of key points in the ground-truth hypothesis might be less than three, so that multiple points can be given. E.g., there's only one key point in the ground-truth hypothesis, and the proposed hypothesis covers the one key point nearly identically, it's possible to give 2 points, 4 points, and 5 points. In this case, we should choose score from 4 points and 5 points, depending on the existence and quality of extra key points. 'Leveraging a key point nearly identically as in the ground-truth hypothesis means that in the proposed hypothesis, the same (or very related) concept (key point) is used in a very similar way with a very similar goal compared to the ground-truth hypothesis. When judging whether an extra key point has apparent flaws, you should use your own knowledge and understanding of that discipline to judge, rather than only relying on the count number of pieces of extra key point to judge. Importantly, we should focus on whether the fundamental key points match, rather than being influenced by how complex, sophisticated, or advanced the proposed methods appear. A hypothesis that introduces high-level techniques or intricate methodologies does not necessarily mean it is a disadvantage with the ground-truth hypothesis. The core concern is whether the essential key points are correctly captured and utilized.Please evaluate the proposed hypothesis based on the ground-truth hypothesis. The proposed hypothesis is: ", "The ground-truth hypothesis is: ", "The key points in the ground-truth hypothesis are: "

\begin{table*}[h]
  \centering
  \renewcommand{\arraystretch}{1.1}
  \resizebox{0.9\columnwidth}{!}{
  \begin{tabular}{l p{13cm}}
    \toprule
    \textbf{Score} & \textbf{Criteria} \\
    \midrule
    5 Points & 
    (1) Covers three key points (or all key points) in the ground-truth hypothesis, with each key point leveraged nearly identically to the ground-truth hypothesis. (2) Does not contain any extra key point that is redundant, unnecessary, unhelpful, or harmful. \\
    \midrule
    4 Points & 
    (1) Covers three key points (or all key points) in the ground-truth hypothesis, with each key point leveraged nearly identically to the ground-truth hypothesis. (2) However, it also contains extra key point(s) that are redundant, unnecessary, unhelpful, or harmful. \\
    \midrule
    3 Points & 
    (1) Covers two key points in the ground-truth hypothesis, with each key point leveraged nearly identically to the ground-truth hypothesis. (2) Does not cover all key points in the ground-truth hypothesis. (3) May or may not contain extra key points. \\
    \midrule
    2 Points & 
    (1) Covers one key point in the ground-truth hypothesis and leverages it nearly identically to the ground-truth hypothesis. (2) Does not cover all key points in the ground-truth hypothesis. (3) May or may not contain extra key points. \\
    \midrule
    1 Point & 
    (1) Covers at least one key point in the ground-truth hypothesis, but all the covered key points are used differently from the ground-truth hypothesis. (2) May or may not contain extra key points. \\
    \midrule
    0 Points & 
    The proposed hypothesis does not cover any key point in the ground-truth hypothesis. \\
    \bottomrule
  \end{tabular}}
  \caption{Scoring criteria for hypothesis evaluation.}
  \label{tab:scoring_criteria}
\end{table*}

\subsection{Prompt for Pairwise Ranking}
\label{appendix_prompt_for_ranking}
You are assisting scientists with their research. Given a research question and two research hypothesis candidates proposed by large language models, your task is to predict which hypothesis is a better research hypothesis. By 'better', we mean the hypothesis is more valid and effective for the research question. 
Please note the following:

(1) Neither hypothesis has been tested experimentally. However, some large language model generated hypothesis might contain expected performance of the hypothesis. Well, just do not believe any of the descriptions of the expected performance or the effect of the hypothesis. Instead, only focus on the technical contents and predict which hypothesis you think will be more effective for the research question if tested in real experiments.

(2) You should remember that, here, we only focus on whether the general direction or major components of the hypothesis are more effective. Providing additional details or making the content more comprehensive is neither an advantage nor a disadvantage. More detailed and multifaceted strategies, additional complexity, and potential challenges are neither advantages nor disadvantages. What truly matters is the fundamental, intrinsic core idea.
The research question is: <the background question of this paper>
Research hypothesis candidate 1 is: <the ground-truth hypothesis>
Research hypothesis candidate 2 is: <the negative hypothesis>

Now, please predict which hypothesis you think will be more effective for the research question if tested in real experiments. 

\subsection{Prompts for Mutate, Refine, and Recombine}
\label{Prompts for mutate, refine, and recombine}
\textbf{Prompt for mutation:} You are helping with the scientific hypotheses generation process. We in general split the period of research hypothesis proposal into three steps. Firstly it's about the research background, including finding a good and specific background research question, and an introduction of the previous methods under the same topic; Secondly its about finding inspirations (mostly from literatures), which combined with the background research question, can lead to a impactful research hypothesis; Finally it's hypothesis generation based on the background research question and found inspirations. Take backpropagation as an example, the research question is how to use data to automatically improve the parameters of a multi-layer logistic regression with data, the inspiration is the chain rule in mathematics, and the research hypothesis is the backpropagation itself. \\nNow we have identified a good research question, an introduction of previous methods, and a core inspiration in a literature for this research question. The experts know that a proper mixture of these components will definitely lead to a valid, novel, and meaningful research hypothesis. In fact, they already have tried to mix them to compose some research hypotheses (that are supposed to be distinct from each other). Please try to explore a new meaningful way to combine the inspiration with the research background to generate a new research hypothesis that is distinct with all the previous hypotheses in terms of their main method. The new research hypothesis should ideally be novel, valid, ideally significant, and be enough specific in its methodology. Please note that here we are trying to explore a new meaningful way to leverage the inspiration along with the previous methods (inside or outside the introduction) to better answer the background research question, therefore the new research hypothesis should try to leverage or contain the key information or the key reasoning process in the inspiration, trying to better address the background research question. It means the new research hypothesis to be generated should at least not be completely irrelevant to the inspiration or background research question. In addition, by generating distinct hypothesis, please do not achieve it by simply introducing new concept(s) into the previous hypothesis to make the difference, but please focus on the difference on the methodology of integrating or leveraging the inspiration to give a better answer to the research question  (in terms of the difference on the methodology, concepts can be introduced or deleted).

\textbf{Prompt for refine:} You are helping with the scientific hypotheses generation process. We in general split the period of research hypothesis proposal into four steps. Firstly it's about finding a good and specific background research question, and an introduction of the previous methods under the same topic; Secondly its about finding inspirations (mostly from literatures), which combined with the background research question, can lead to a impactful research hypothesis; Thirdly it's about finding extra knowledge that work along with the inspiration can lead to a more complete hypothesis. Finally it's hypothesis generation based on the background research question, the found inspirations, and the extra knowledge. Now we have identified a good research question, a core inspiration in a literature for this research question, and extra knowledge. With them, we have already generated a preliminary research hypothesis. We have also obtain feedbacks on the hypothesis from domain experts in terms of novalty, validity, significance, and clarity. With these feedbacks, please try your best to refine the hypothesis. Please note that during refinement, do not improve a hypothesis's significance by adding expectation of the performance gain of the method or adding description of its potential impact, but you should work on improving the method itself (e.g., by adding or changing details of the methodology).

\textbf{Prompt for recombine:} You are helping with the scientific hypotheses generation process. We in general split the period of research hypothesis proposal into three steps. Firstly it's about the research background, including finding a good and specific background research question, and an introduction of the previous methods under the same topic; Secondly its about finding inspirations (mostly from literatures), which combined with the background research question, can lead to a impactful research hypothesis; Finally it's hypothesis generation based on the background research question and found inspirations. Now we have identified a good research question, an introduction of previous methods, and a core inspiration in a literature for this research question. In addition, several experts have already come out of several different hypotheses on how to leverage the inspiration to generate a novel, valid, and significant research hypothesis for the background research question. Please find the bright parts in these hypotheses, leverage the bright parts from them,  modify and combine the good parts of them to generate a better research hypothesis in terms of clarity, novelty, validness, and significance (ideally than any of the given hypotheses). It is not necessary to include methods from every given hypothesis, especially when it is not a good hypothesis. But in general you should try your best to benefit from every given hypothesis. In fact, a researcher has already tried to propose hypothesis based on these information, and we have obtained the feedback to his hypothesis, from another respectful researcher. Please try to leverage the feedback to improve the hypothesis, you can leverage all these provided information as your reference.



\subsection{Detailed Example of Framework}
\label{sec:example_pipeline}

To provide a concrete understanding of the LLM Based Agentic Framework operation, we present a complete example of the actual inputs and outputs from our pipeline below. This example shows exactly what the LLM reads and generates at each step.

\noindent\textbf{Input Target Paper} \\
\textbf{Title:} The tunable absorber films of grating structure of AlCuFe quasicrystal with high Q and refractive index sensitivity \\
\textbf{Abstract:} Due to the continuous development of terahertz field and materials science, terahertz absorbers have received extensive attention. The paper proposes a tunable absorber film of grating structure of AlCuFe quasicrystal. The absorber film composed of a AlCuFe quasicrystal grating structure and a gold substrate, making it easy to process. The absorber film has three perfect absorption bands in its working frequency band... (omitted for brevity) ... In general, the absorber film has great application prospects in fields such as biomedicine and communication science.

\noindent\textbf{Extracted Research Question} \\
How can we design a tunable, high-performance terahertz absorber that is easy to fabricate, offers multi-band absorption capabilities, and exhibits high sensitivity for sensing applications?

\noindent\textbf{Extracted Background Survey} \\
Certainly! Here’s a more concise background survey without any links: To provide a background survey relevant to the design of tunable, high-performance terahertz absorbers, the following points summarize the pertinent knowledge: 
1. **Terahertz Frequency Range**: The terahertz (THz) frequency range, spanning from 0.1 to 10 THz, is essential for applications in imaging, spectroscopy, and communication. Many materials exhibit unique spectral characteristics within this range, highlighting the need for effective absorbers.
2. **Metamaterials**: Metamaterials are engineered materials that demonstrate properties not found in nature, enabling novel control of electromagnetic waves. The development of metamaterial absorbers has advanced significantly, achieving high absorption efficiencies through innovative structural designs.
3. **Material Properties**: The performance of terahertz absorbers is heavily influenced by the choice of materials. High conductivity, tunable optical properties, and appropriate structural configurations are critical for achieving desired electromagnetic responses.
4. **Absorption Mechanisms**: Key mechanisms of electromagnetic wave absorption include guided mode resonance and localized surface plasmon resonance (LSPR). Guided mode resonance enhances electromagnetic waves within structured materials, while LSPR involves the oscillation of surface electrons in metallic materials, leading to increased absorption.
This background survey captures the essential concepts and developments leading up to the current research, situating the inquiry within the broader context of terahertz absorber technology and metamaterial design.

\noindent\textbf{Inspiration Decomposition}

\noindent\textit{Inspiration 1}

\textbf{Title:} PtTe2-based type-II dirac semimetal and its van der waals heterostructure for sensitive room temperature terahertz photodetection

\textbf{Relation:} Based on the abstract provided, the paper titled "PtTe2-based type-II dirac semimetal and its van der waals heterostructure for sensitive room temperature terahertz photodetection" appears to be relevant to the inspiration for using Dirac semimetals for their tunable optical properties and strong response in the terahertz range. The paper discusses the use of PtTe2, a type-II Dirac semimetal, for terahertz photodetection, highlighting its strong photoresponsivity and fast response time at room temperature. This aligns well with the inspiration of leveraging Dirac semimetals for tunable and high-sensitivity applications in the terahertz range. Therefore, this paper can serve as a suitable inspiration for the use of Dirac semimetals in the proposed absorber film.

\textbf{Analyze:} The paper explores the potential of PtTe2, a Dirac semimetal, in terahertz applications, focusing on its strong optical response and high performance in photodetection, which aligns with the inspiration of using Dirac semimetals for their tunable properties and strong terahertz response.

\noindent\textit{Inspiration 2}

\textbf{Title:} Dual-band tunable narrowband near infrared light trapping control based on a hybrid grating-based Fabry Perot structure

\textbf{Relation:} Based on the abstract provided, the paper titled "Dual-band tunable narrowband near infrared light trapping control based on a hybrid grating-based Fabry Perot structure" appears to be relevant to the inspiration for adopting a grating structure to facilitate easy fabrication and to enable multi-band absorption. The paper discusses the use of a hybrid grating-based Fabry-Perot structure for light manipulation, demonstrating how it can achieve dual-band absorption with tunable spectral selectivity. This aligns with the inspiration for using a grating structure to achieve multi-band absorption in the proposed absorber film. Therefore, this paper can serve as a suitable inspiration for the use of a grating structure in the proposed absorber film.

\textbf{Analyze:} The paper explores a hybrid grating-based Fabry-Perot structure for near-infrared light trapping, showcasing its ability to achieve dual-band absorption and tunability, which aligns with the inspiration of using a grating structure for easy fabrication and multi-band absorption.

\noindent\textit{Inspiration 3}

\textbf{Title:} Perfect metamaterial absorber

\textbf{Relation:} Based on the abstract provided, the paper titled "Perfect metamaterial absorber" is directly relevant to the inspiration for achieving perfect absorption using metamaterial absorbers. The paper presents a design for a metamaterial absorber with near-unity absorbance, which aligns with the goal of achieving high absorption efficiency in the proposed absorber film. This foundational work on metamaterial absorbers provides the theoretical underpinning for the design approach discussed in the paper. Therefore, this paper can serve as a suitable inspiration for the concept of achieving perfect absorption through metamaterial absorbers.

\textbf{Analyze:} The paper presents a design and demonstration of a metamaterial absorber with near-unity absorbance, offering a theoretical and practical foundation for achieving high absorption efficiency, which aligns with the inspiration for using metamaterial absorbers in the proposed design.

\noindent\textbf{Extracted Hypothesis} \\
A tunable absorber film consisting of an AlCuFe quasicrystal (a Dirac semimetal) grating structure on a gold substrate can achieve perfect multi-band absorption in the terahertz range, with tunability provided by variations in the Fermi energy of the Dirac semimetal, grating structure parameters, or incident polarization. The absorption mechanism is attributed to guided mode resonance within the grating structure and local surface plasmon resonance (LSPR) on the AlCuFe surface, resulting in high Q-factor and high refractive index sensitivity suitable for sensing applications.

\noindent\textbf{Extracted Experimental Design} \\
Thank you for your patience, and I appreciate your guidance throughout this process. Upon reviewing the previous summarization of the experiments, I have ensured to maintain the original structure while adding necessary details for clarity and completeness. Below is the refined summary with detailed methodologies for each experiment, ensuring that no important aspects are overlooked:

\#\#\# 1. Absorption Spectrum Measurement
- **Objective**: Measure the absorption spectrum of the absorber film under different polarization modes (TE and TM waves) to determine its absorption characteristics.
- **Method**:
  - **Setup**: Utilize a terahertz spectroscopy system equipped with a tunable terahertz source and a detector.
  - **Sample Preparation**: Fabricate the absorber film with a grating structure of AlCuFe quasicrystal on a gold substrate, ensuring the dimensions correspond to the parameters defined in the study.
  - **Procedure**:
    - Position the sample in the optical path and ensure proper alignment.
    - Configure the source to emit terahertz waves over a frequency range (e.g., 0.1 to 10 THz).
    - Measure and record the transmission and reflection for TE and TM polarizations separately.
    - Use the formula $A = 1 - R - T$ to calculate the absorptivity, where $R$ is measured reflectivity and $T$ is measured transmittance.
  - **Expected Outcomes**: Identify three distinct perfect absorption bands and observe the transition to a single absorption band when the polarization direction changes, confirming the tunability of the absorber.

\#\#\# 2. Impedance Matching Analysis
- **Objective**: Calculate the relative impedance of the absorber film to verify its effectiveness in absorption.
- **Method**:
  - **Setup**: Employ a vector network analyzer (VNA) to measure S-parameters of the absorber film.
  - **Procedure**:
    - Connect the sample to the VNA and perform calibration.
    - Measure the S11 (reflection coefficient) and S21 (transmission coefficient) parameters across the defined frequency range.
    - Calculate the relative impedance $Z$ using:
      \[ Z = \frac{(1 + S_{11})}{(1 - S_{11})} \cdot Z_0 \]
      where $Z_0 \approx 376.7 \, \Omega$.
  - **Expected Outcomes**: Determine the frequency ranges where the relative impedance closely approaches the free space impedance, validating the theoretical basis for effective absorption.

\#\#\# 3. Electric Field Distribution and Absorption Mechanism Analysis
- **Objective**: Analyze the electric field distribution and identify the absorption mechanisms (guided mode resonance and LSPR).
- **Method**:
  - **Setup**: Use simulation software such as CST Microwave Studio.
  - **Procedure**:
    - Model the absorber film in the software with the specified geometry and material properties.
    - Set the simulation to observe electric field distribution under normal incident waves.
    - Simulate the electric field for different modes of the absorber film, capturing data for analysis.
    - Visualize the electric field patterns to identify areas of strong enhancement.
  - **Expected Outcomes**: Confirm the mechanisms of absorption, demonstrating how structural parameters influence performance.

\#\#\# 4. Parameter Variation and Physical Tuning Experiments
- **Objective**: Investigate how varying physical parameters (lengths of grating columns) affect absorption.
- **Method**:
  - **Setup**: Use both terahertz spectroscopy and simulation software.
  - **Procedure**:
    - Systematically modify parameters such as the lengths $D_1$, $D_2$, $D_3$ and widths $L_1$, $L_2$ of the grating structure.
    - Measure and record the absorption spectra for each configuration, ensuring all other parameters are held constant.
    - Analyze the...

\subsection{Examples of Inspiration Retrieval, Hypothesis Composition and Hypothesis Ranking}
\label{sec:subtask_examples}

We provide a example for each of the three subtasks below.

\noindent\textbf{1. Inspiration Retrieval}

In this subtask, the model receives the research question, the instruction, and a set of candidate papers (15 candidates per round).

\noindent\textbf{Research Question:} How can a low-cost, eco-friendly adsorbent be developed to efficiently remove methylene blue (MB), a common pollutant, from aqueous solutions?

\noindent\textbf{GPT-4o Selected Inspirations:} \\
\noindent\textbf{Title:} Biochar-supported zerovalent iron for removal of various contaminants from aqueous solutions. \\
\textbf{Reason:} This study could serve as an inspiration because it addresses the enhancement of adsorption and stability properties through the integration of biochar and zero-valent iron (ZVI), which is both cost-effective and environmentally friendly. The functional modifications to biochar for improving sorption capacity and its suitability for methylene blue (MB) removal are directly applicable to the research question. Additionally, the study emphasizes the design of composite materials for efficient sorption, which aligns with the objective of developing a low-cost, eco-friendly adsorbent.

\noindent\textbf{Title:} Uniform silver nanowires synthesis by reducing AgNO$_3$ with ethylene glycol in the presence of seeds and poly(vinyl pyrrolidone). \\
\textbf{Reason:} While this paper primarily discusses the synthesis of silver nanowires, its focus on controlling particle characteristics (e.g., high aspect ratio and uniformity) using cost-effective and tunable methods could inform the structural optimization of adsorbents for methylene blue removal. The surface morphology and characteristics of the nanowires could provide insights into creating porous structures or functionalized surfaces to enhance MB adsorption efficiency.

\noindent\textbf{Title:} Properties and Possibilities of Using Biochar Composites Made on the Basis of Biomass and Waste Residues Ferryferrohydrosol Sorbent \\
\textbf{Reason:} This study is relevant as it explores the use of biochar composites derived from biomass and enriched with metals. The composite materials' enhanced potential for MB adsorption, combined with their eco-friendly and cost-efficient production methods, directly addresses the need for low-cost and environmentally sustainable adsorbents. The detailed evaluation of functional groups and adsorption mechanisms could guide the development of efficient MB removal solutions.

\textit{Among these, Inspirations 1 and 2 are ground-truth inspirations for this task.}

\noindent\textbf{2. Hypothesis Composition}

In this subtask, the model receives the research question and ground-truth inspirations. The goal is to generate a hypothesis and compare it with the ground truth.

\noindent\textbf{Paper:} Dacarbazine-primed carbon quantum dots coated with breast cancer cell-derived exosomes for improved breast cancer therapy \\
\noindent\textbf{Research Question:} How can we enhance the targeted delivery and therapeutic efficacy of dacarbazine in breast cancer cells while minimizing systemic toxicity?

\noindent\textbf{GPT-4o Generated Hypothesis:}
The development of a hybrid delivery system that incorporates sugar-originated carbon nanodots (sCND) conjugated with dacarbazine and encapsulated within a glucose-mimetic hydrogel featuring pH-responsive properties and a dynamic covalent ketone-hydrazone bonding network will improve the targeted delivery and therapeutic efficacy of dacarbazine in GLUT1-overexpressing breast cancer cells. This system is designed to leverage the acidic tumor microenvironment to achieve concerted competitive inhibition of GLUT1, precise release of dacarbazine, and enhanced retention of therapeutic agents at tumor sites, while significantly reducing systemic toxicity by avoiding off-target drug dissemination to healthy tissues.

\noindent\textbf{Ground-truth Hypothesis:}
Dacarbazine-primed carbon quantum dots coated with breast cancer cell-derived exosomes (Ex-DC@CQDs) can provide enhanced targeted delivery to breast cancer cells, leading to improved drug accumulation at tumor sites, increased therapeutic efficacy, and reduced systemic toxicity.

\noindent\textbf{Score:} 2 points (according to Table \ref{tab:scoring_criteria})

\noindent\textbf{Reason:} The proposed hypothesis incorporates the use of carbon nanodots (sCNDs) conjugated with dacarbazine, which aligns with the second key point of the groundtruth hypothesis regarding the integration of carbon quantum dots (CQDs) with chemotherapeutic agents. However, the proposed hypothesis does not utilize exosomes for targeted drug delivery as described in the first key point of the groundtruth hypothesis. Instead, it employs a glucose-mimetic hydrogel with pH-responsive properties and dynamic covalent ketone-hydrazone bonding, targeting GLUT1-overexpressing cells and leveraging the acidic tumor microenvironment. While the goal of enhancing delivery and reducing systemic toxicity is similar to the groundtruth hypothesis, the proposed mechanism is significantly different, focusing on chemical and biological properties rather than relying on exosome-mediated targeting. The proposed hypothesis contains several extra components, such as the glucose-mimetic hydrogel, pH-responsive properties, and a dynamic covalent ketone-hydrazone bonding network, which are not in groundtruth hypothesis. However, these extra points can be helpful for the research purpose.

\noindent\textbf{3. Hypothesis Ranking}

In this subtask, the model compares the ground-truth hypothesis with multiple negative hypotheses using pairwise evaluations, with order reversal to mitigate position bias.

\noindent\textbf{Research Question:} How does the relationship between supermassive black holes (SMBHs) and their host galaxies evolve over time, especially during the star-formation and black-hole-growth peak at redshifts 1--3?

\noindent\textbf{Ground-truth Hypothesis:} During the peak epoch ($z \approx 2$), some host galaxies grow faster than their SMBHs; evidence includes the detection of an undermassive SMBH accreting at a super-Eddington rate in a high-redshift quasar.

\noindent\textbf{Negative Hypothesis:} At redshifts 1--3, the coevolution of supermassive black holes (SMBHs) and their host galaxies is driven by interactions with companion galaxies and cold-gas inflows, which shape molecular-gas accretion and star formation under AGN feedback. Using high-resolution ALMA data, we will quantify how companions and gas properties affect gas consumption, star-formation efficiency, and host-galaxy dynamics.

\noindent\textbf{Ranking Result:} Given the research question, GPT-4o incorrectly selected the negative hypothesis as the better one, illustrating the difficulty of the ranking task.

\end{document}